\documentclass[10pt,twocolumn,letterpaper]{article}

\makeatletter
\def\input@path{{required_style_files/}{required_class_files/}}
\makeatother

\usepackage[pagenumbers]{cvpr}

\usepackage{amsmath}
\usepackage{amssymb}
\usepackage{booktabs}
\usepackage{graphicx}
\usepackage{microtype}
\usepackage{xcolor}

\definecolor{cvprblue}{rgb}{0.21,0.49,0.74}
\usepackage[breaklinks,colorlinks,allcolors=cvprblue]{hyperref}

\def\paperID{}
\def\confName{CVPR}
\def\confYear{}

\title{Think Sparse, Predict Dense:\\
Continuous Thought Machines for Image Super-Resolution}

\author{
Zekai Shi \\
Xi'an Jiaotong University
}

\begin{document}
\maketitle

\begin{abstract}
Continuous Thought Machines introduce an internal temporal dimension in which neuron-level histories and synchronization-derived representations evolve over a sequence of thought ticks. Extending this mechanism to dense visual prediction is non-trivial, because tasks such as image super-resolution require spatial evidence to remain available at every output location rather than being compressed into a single global representation. In the proposed window-level use of CTM, the thought dynamics produce a compact summary representation for each local window. DQ-CTM transforms this compact thought representation into window-aligned dense queries through a structured low-rank, parameter-efficient compact-to-dense query mechanism. Each position within a window receives its own query, while shared thought dynamics progressively refine the dense representation across ticks. In its super-resolution instantiation, termed ThinkSR, encoded feature maps are partitioned into local visual windows without token pooling, restored to the original feature field after shared refinement, and decoded into a high-resolution image. Preliminary experiments under a fixed four-tick training horizon reveal a progressive reconstruction trajectory. PSNR-Y increases from 28.1045 dB at $T=0$ to 30.2817 dB at $T=4$, while PSNR-RGB increases from 26.6271 dB to 28.7781 dB and the mean $\ell_1$ error decreases from 0.034602 to 0.023545. All 100 evaluated images improve from $T=1$ to $T=4$. These initial results establish the feasibility of sparse latent thought for dense spatial reconstruction and motivate broader continuous-thought architectures for dense vision.
\end{abstract}


\section{Introduction}
\label{sec:introduction}

Single-image super-resolution (SR) is commonly modeled as a fixed-depth mapping from a low-resolution observation to a high-resolution estimate. Modern convolutional and Transformer-based systems can represent this mapping with substantial capacity, from residual reconstruction networks such as EDSR~\cite{lim2017edsr} to shifted-window restoration models such as SwinIR~\cite{liang2021swinir}. Despite their architectural differences, the usual inference procedure remains a single feed-forward pass whose depth is selected before any image is observed. The same amount of computation is therefore allocated to an easy smooth region, a regular edge, and a severely aliased texture. This fixed allocation is practical and predictable, but it leaves little room to expose an internal reconstruction trajectory or to study how an estimate changes as the model continues processing. Recursive SR models have shown that parameter sharing across depth is a useful alternative~\cite{kim2016drcn}; however, recursion alone does not define a compact state whose temporal evolution can coordinate a dense field of local predictions. A useful iterative formulation should preserve spatial evidence while making the evolving computation measurable at each step.

The Continuous Thought Machine (CTM) introduces an internal thought dimension in which computation unfolds over discrete ticks~\cite{darlow2025ctm}. Its defining ingredients are neuron-level temporal processing over pre-activation histories and a synchronization-derived representation constructed from interactions among neural activities. This differs from treating depth merely as a stack of independently parameterized layers and from standard self-attention blocks~\cite{vaswani2017attention}. The compact activated state can change as histories accumulate, while synchronization provides a representation through which the model reads or organizes information. Existing CTM demonstrations naturally emphasize compact queries and low-dimensional decisions, such as classification, action selection, or the endpoint of a reasoning process. Dense prediction presents a different interface. The output cannot be represented only by a class token or a small decision vector: the computation must remain coupled to a spatial carrier whose locations retain the evidence required for reconstruction. The central issue is therefore not simply whether CTM dynamics can process images, but how their compact temporal state can influence a large, location-indexed representation throughout thought.

This report asks: \emph{How can compact continuous thought govern dense visual prediction without collapsing the underlying spatial evidence?} The resulting difficulty is termed the \emph{sparse-thought--dense-output mismatch}. A thought process may be compact in its activated latent state and synchronization statistics, yet a dense vision task demands a prediction at every output coordinate. SR is a strict instance of this mismatch because its target grid is spatially denser than its input grid, all high-resolution locations require estimates, and lost local evidence directly appears as blur, ringing, or fabricated texture. Pooling all visual tokens into a single global descriptor would make the thought interface compact, but it would also remove the explicit correspondence needed to recover fine structure. Conversely, retaining a dense token field without a defined interaction with the temporal state would reduce continuous thought to an auxiliary recurrent signal. The desired construction must therefore separate two roles: a persistent carrier stores location-specific visual content, while a compact thought process reads from and modulates that carrier without changing its token cardinality.

The Dense-Query Continuous Thought Machine (DQ-CTM) instantiates this separation, and its SR realization is termed ThinkSR. Within each local window, CTM processing produces a compact thought summary. DQ-CTM converts that summary into window-aligned dense queries through structured low-rank query generation, assigning a distinct query to every window position without separately parameterizing each query. A SwinIR-style encoder produces a dense feature field, which is partitioned into non-overlapping local windows while preserving every token in each window. Shared continuous-thought dynamics maintain a compact activated state and temporal history. Synchronization-derived thought representations query or modulate the window tokens through a dense, location-preserving update, after which the windows are restored to their original feature layout and decoded into an image. Because the thought parameters are shared across ticks, the model exposes a progressive sequence of reconstructions within the learned horizon rather than a set of separately parameterized predictors. The present V1 focuses on whether this interface is feasible and measurable; it does not claim adaptive stopping or reliable extrapolation beyond the trained horizon. Its contributions are:
\begin{itemize}
    \item The sparse-thought--dense-output mismatch is formulated as a fundamental challenge when extending continuous thought from compact visual decisions to dense prediction.
    \item A Dense-Query Continuous Thought Machine is introduced as a low-rank, parameter-efficient compact-to-dense query mechanism that maps each compact window summary to distinct, window-aligned queries while retaining the persistent dense visual carrier.
    \item The formulation is instantiated in image super-resolution, where preliminary results demonstrate a progressively improving reconstruction trajectory within the learned thought horizon.
\end{itemize}

\begin{figure*}[t]
    \centering
    \includegraphics[width=\textwidth]{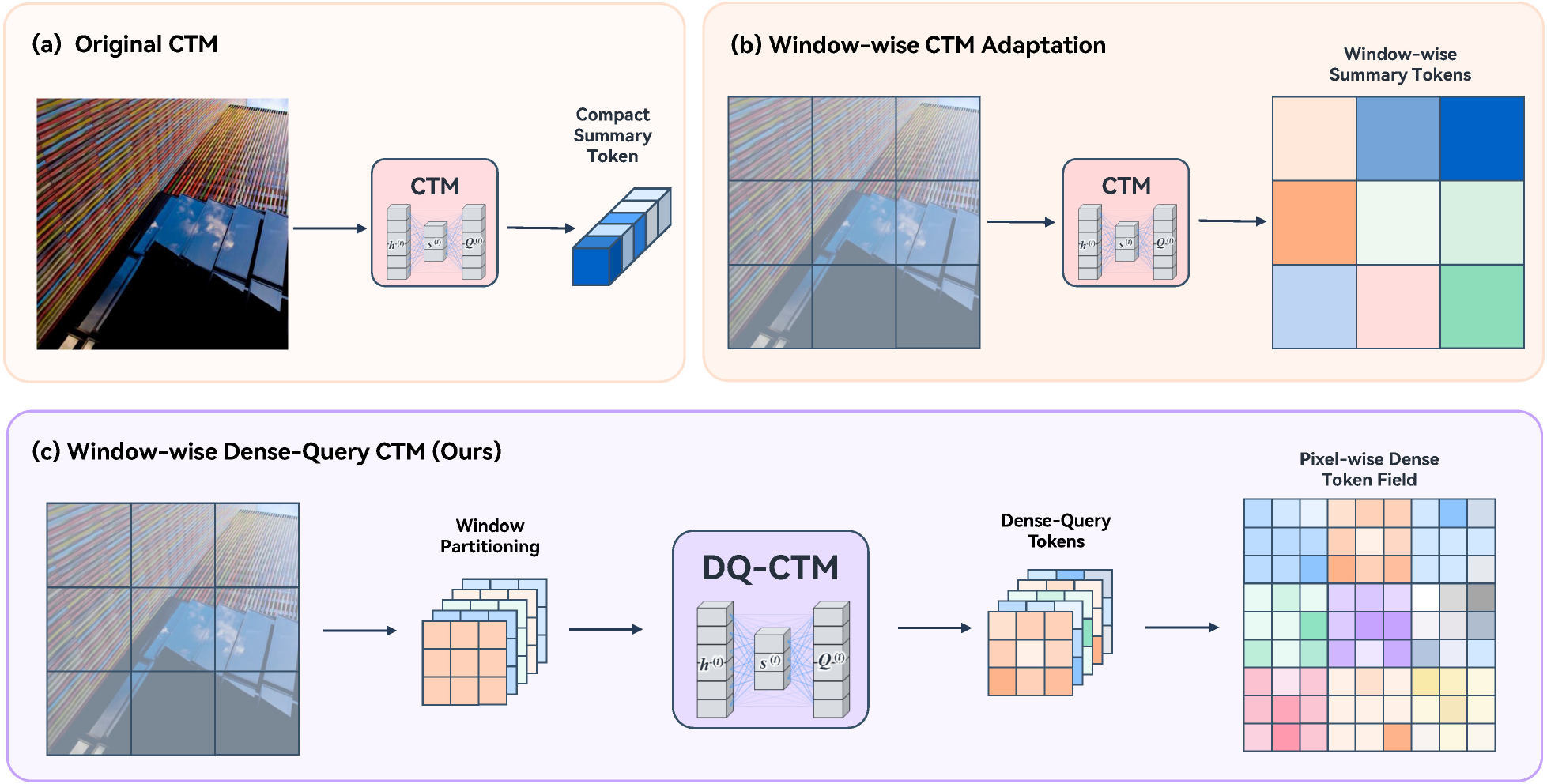}
    \caption{\textbf{From compact CTM representations to window-aligned dense queries.} The original CTM compresses visual information into a compact abstract token. A direct window-wise adaptation preserves coarse window locality but still produces only one summarized token for each window, collapsing the spatial identities within that window. DQ-CTM instead produces window-aligned dense query tokens, providing a position-wise representation compatible with dense visual prediction.}
    \label{fig:ctm_comparison}
\end{figure*}

The compact output formulation of the original CTM is naturally suited to image-level reasoning, where a small number of abstract representations can support a final decision. Directly applying the same mechanism to individual visual windows retains coarse locality, but each window is still reduced to a single summarized token. Such a window-level abstraction does not natively satisfy the position-aligned output requirement of dense prediction tasks, because spatial identities within each window remain collapsed. DQ-CTM addresses this mismatch by transforming the compact thought representation into window-aligned dense queries, allowing recurrent reasoning to operate without discarding the per-position token structure. This progression is summarized in Fig.~\ref{fig:ctm_comparison}.

\section{Dense Continuous Thought for Image Super-Resolution}
\label{sec:method}

\subsection{Problem Formulation}

Let $\mathbf{I}_{\mathrm{LR}}$ denote a low-resolution image and $\mathbf{I}_{\mathrm{HR}}$ its high-resolution target at scale factor $s$. An encoder maps the observation to a dense feature field $\mathbf{F}$. Computation then evolves over thought ticks $t\in\{0,\ldots,T\}$, producing an image estimate $\widehat{\mathbf{I}}^{(t)}$ at each tick. The objective is to preserve the spatial evidence in $\mathbf{F}$ while allowing a compact temporal process to refine the estimate. The V1 specification intentionally states this interface without publishing implementation-level intermediate dimensions.

\subsection{Persistent Dense Visual Carrier}

A SwinIR-style encoder~\cite{liang2021swinir} extracts the dense feature field. A spatial window operator $\mathcal{W}$ partitions the feature map into non-overlapping local windows:
\begin{equation}
    \mathbf{F}=\mathcal{E}(\mathbf{I}_{\mathrm{LR}}),
    \qquad
    \mathbf{X}=\mathcal{W}(\mathbf{F}).
    \label{eq:carrier}
\end{equation}
Each window retains its complete token sequence. No token pooling is applied, and the CTM-facing input and output contain the same number of visual tokens. The carrier is called persistent because location-indexed content remains available across all thought ticks, even though the compact state used to govern its updates evolves temporally.

\subsection{Dense-Query Continuous Thought}

At tick $t$, the thought module maintains a compact activated state $\mathbf{h}^{(t)}$ together with a temporal history of neuron-level pre-activations. A neuron-level model processes this history and produces a synchronization-derived thought representation $\mathbf{s}^{(t)}$. In the proposed window-level configuration, these dynamics yield a compact summary that is transformed into window-aligned dense queries by a low-rank dense query projection. Every position in a window receives a distinct query, and the same parameter-efficient query mechanism is reused across shared thought ticks. Rather than replacing the dense carrier, synchronization is used to read from or modulate its tokens. The interface can be summarized abstractly as
\begin{align}
\mathbf{h}^{(t+1)}
&=\mathcal{T}_{\theta}\!\left(
\mathbf{h}^{(t)},
\operatorname{Read}(\mathbf{X}^{(t)},\mathbf{s}^{(t)})
\right),
\label{eq:thought}\\
\mathbf{X}^{(t+1)}
&=\mathcal{U}_{\theta}\!\left(
\mathbf{X}^{(t)},\mathbf{s}^{(t)}
\right).
\label{eq:update}
\end{align}
Here $\mathcal{T}_{\theta}$ denotes the compact thought transition and $\mathcal{U}_{\theta}$ a dense, token-count-preserving update. Parameters $\theta$ are shared across ticks. Equations~\eqref{eq:thought}--\eqref{eq:update} specify the intended interface rather than every operation of the original CTM. The V1 report characterizes the compact-to-dense transformation only by this functional role; its factorization, intermediate dimensions, coefficient construction, and implementation-specific update order are intentionally omitted.

\begin{figure*}[t]
    \centering
    \includegraphics[width=\textwidth]{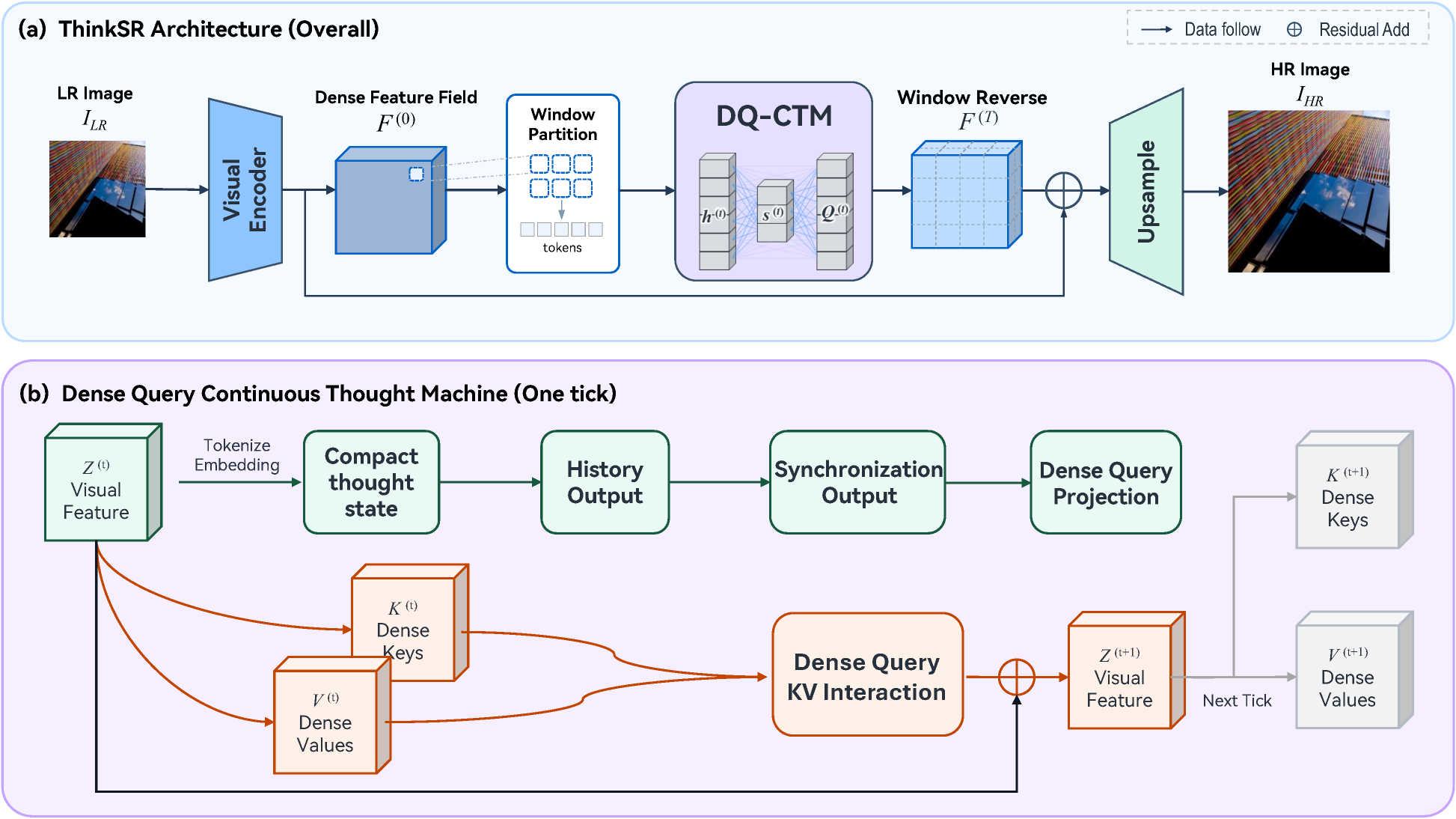}
    \caption{\textbf{Overall architecture of ThinkSR with DQ-CTM.} A dense visual encoder extracts a spatial feature field, which is partitioned into local windows and recurrently refined by the shared DQ-CTM module. The updated window tokens are restored to the spatial feature field and decoded into the high-resolution output. Intermediate predictions illustrate progressive reconstruction across thought ticks.}
    \label{fig:thinksr_architecture}
\end{figure*}

Figure~\ref{fig:thinksr_architecture} summarizes the complete ThinkSR pathway from dense visual encoding and window partition through shared DQ-CTM ticks, window reversal, dense reconstruction, and progressive outputs.

\subsection{Dense Reconstruction}

The updated windows are restored to their spatial arrangement and passed to an SR decoder $\mathcal{D}$:
\begin{equation}
\widehat{\mathbf{I}}^{(t)}=
\mathcal{D}\!\left(\mathcal{W}^{-1}(\mathbf{X}^{(t)})\right).
\label{eq:decode}
\end{equation}
The encoder, thought transition, dense update, and decoder parameters are shared at every thought tick; increasing $t$ within the trained unroll does not introduce a separately parameterized reconstruction head.

\subsection{Training Objective}

The current V1 experiment record describes supervision at the final thought tick $T=4$ with a pixel reconstruction objective. Using the reported $\ell_1$ objective, the training loss is
\begin{equation}
\mathcal{L}_{\mathrm{rec}}=
\left\|\widehat{\mathbf{I}}^{(4)}-\mathbf{I}_{\mathrm{HR}}\right\|_1.
\label{eq:loss}
\end{equation}
The current V1 model is optimized using the reconstruction produced at the final supervised thought tick. No per-tick supervision or monotonic-improvement loss is assumed.

\section{Preliminary Experiments}
\label{sec:experiments}

\subsection{Experimental Setting}

The preliminary $\times4$ SR model is trained on the 800-image DIV2K training set using $48\times48$ LR patches and corresponding $192\times192$ HR patches. It has 1.13M parameters, uses Adam, and is unrolled with a fixed training thought horizon $T_{\mathrm{train}}=4$. The reported thought sweep evaluates 100 DIV2K validation images at $T=0,\ldots,4$.

\subsection{Comparison with Lightweight SR Methods}

\begin{table*}[t]
\centering
\caption{Quantitative comparison with representative lightweight image super-resolution methods for $\times4$ reconstruction.}
\label{tab:main_lightweight_sr}
\small
\setlength{\tabcolsep}{2.3pt}
\begin{tabular*}{\textwidth}{@{\extracolsep{\fill}}lccccccc@{}}
\toprule
Method & Params & Set5 & Set14 & BSD100 & Urban100 & Manga109 & Average \\
\midrule
CARN~\cite{ahn2018carn}
& 1.592M
& 32.13/0.8937
& 28.60/0.7806
& 27.58/0.7349
& 26.07/0.7837
& 30.47/0.9084
& 28.970/0.8203 \\
IMDN~\cite{hui2019imdn}
& 0.715M
& 32.21/0.8948
& 28.58/0.7811
& 27.56/0.7353
& 26.04/0.7838
& 30.45/0.9075
& 28.968/0.8205 \\
RFDN~\cite{liu2020rfdn}
& 0.550M
& 32.24/0.8952
& 28.61/0.7819
& 27.57/0.7360
& 26.11/0.7858
& 30.58/0.9089
& 29.022/0.8216 \\
SwinIR-light~\cite{liang2021swinir}
& 0.930M
& 32.44/0.8976
& 28.77/0.7858
& 27.69/0.7406
& 26.47/0.7980
& 30.92/0.9151
& 29.258/0.8274 \\
SRFormer-light~\cite{zhou2023srformer}
& 0.873M
& 32.51/0.8988
& 28.82/0.7872
& 27.73/0.7422
& 26.67/0.8032
& 31.17/0.9165
& 29.380/0.8296 \\
MambaIRv2-light~\cite{guo2025mambairv2}
& 0.790M
& 32.51/0.8992
& 28.84/0.7878
& 27.75/0.7426
& 26.82/0.8079
& 31.24/0.9182
& 29.432/0.8311 \\
CATANet~\cite{liu2025catanet}
& \textbf{0.535M}
& \textbf{32.58/0.8998}
& \textbf{28.90/0.7880}
& \textbf{27.75/0.7427}
& \textbf{26.87/0.8081}
& \textbf{31.31/0.9183}
& \textbf{29.482/0.8314} \\
\midrule
\textbf{DQ-CTM-SR (V1, $T=4$)}
& 1.129M
& 32.1640/0.8954
& 28.6222/0.7826
& 27.6298/0.7377
& 26.1034/0.7859
& 30.3948/0.9085
& 28.983/0.8220 \\
\bottomrule
\end{tabular*}
\vspace{2pt}

\begin{minipage}{\textwidth}
\footnotesize
\textit{Note.} PSNR/SSIM values are evaluated on the luminance channel. Results for competing methods are taken from their original publications, whereas DQ-CTM-SR denotes the directly trained $\times4$ V1 model evaluated using four thought steps. The final column is the unweighted mean over the five datasets.
\end{minipage}
\end{table*}

Table~\ref{tab:main_lightweight_sr} spans established lightweight CNN baselines, mature lightweight Transformer models, and recent lightweight architectures. The V1 model is competitive with established lightweight CNN baselines, while a clear gap remains relative to recent state-of-the-art lightweight architectures, particularly on structure-rich Urban100 and Manga109. This comparison is intended to locate the current V1 performance rather than claim state-of-the-art reconstruction quality.

\subsection{Progressive Reconstruction}

\begin{table}[t]
\centering
\caption{Preliminary reconstruction performance across internal thought ticks under a model trained with a four-tick thought horizon.}
\label{tab:thought-sweep}
\small
\setlength{\tabcolsep}{3.8pt}
\begin{tabular}{rccccc}
\hline
T & L1 & PSNR$_{\mathrm{RGB}}$ & SSIM$_{\mathrm{RGB}}$ & PSNR$_{\mathrm{Y}}$ & SSIM$_{\mathrm{Y}}$ \\
\hline
0 & 0.0386 & 25.45 & 0.7301 & 28.91 & 0.7903 \\
1 & 0.0325 & 26.75 & 0.7822 & 29.88 & 0.8293 \\
2 & 0.0285 & 27.72 & 0.8001 & 30.18 & 0.8360 \\
3 & 0.0249 & 28.54 & 0.8123 & 30.38 & 0.8382 \\
4 & 0.0229 & 28.98 & 0.8193 & 30.47 & 0.8385 \\
\hline
\end{tabular}
\end{table}

Table~\ref{tab:thought-sweep} shows continuous improvement from $T=0$ through $T=4$ in all reported aggregate metrics. From $T=1$ to $T=4$, PSNR-Y increases by 1.4611 dB, and all 100 evaluated images improve over the same interval. The tick-wise gains become smaller near the end of the unroll: the PSNR-Y increment decreases from 0.6928 dB between $T=1$ and $T=2$ to 0.2025 dB between $T=3$ and $T=4$. These observations support progressive refinement only within the learned four-tick horizon. They do not establish comparison with state-of-the-art SR systems or improvement under an indefinitely extended test-time computation budget.

\begin{figure}[t]
    \centering
    \includegraphics[width=\linewidth]{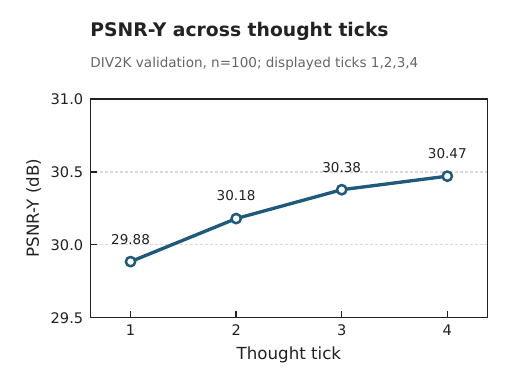}
    \caption{Preliminary reconstruction quality across internal thought ticks under a model trained with a four-tick thought horizon.}
    \label{fig:thought_quality_curve}
\end{figure}

The aggregate PSNR-Y trajectory is visualized in Fig.~\ref{fig:thought_quality_curve}. It covers only the supervised evaluation range $T=0$--$4$ and does not imply improvement beyond the learned horizon.

\begin{figure*}[t]
    \centering
    \includegraphics[width=\textwidth]{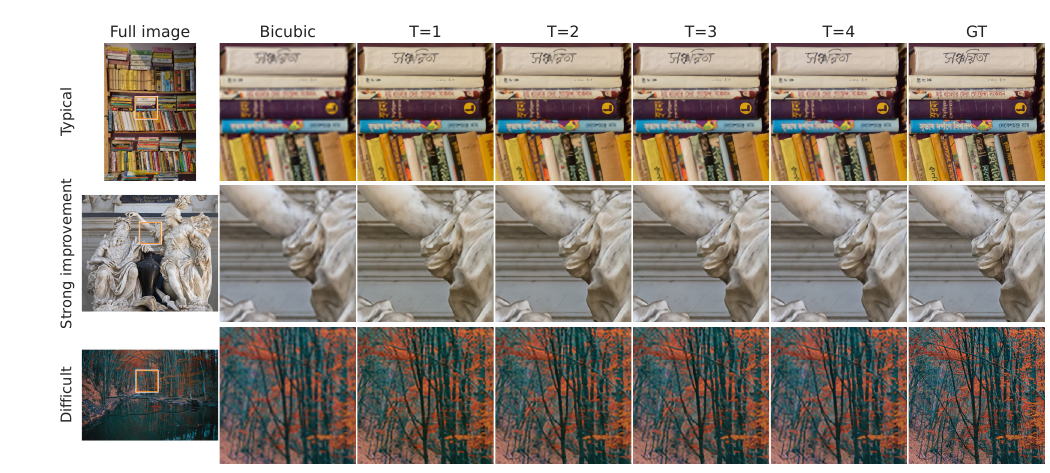}
    \caption{Progressive reconstruction trajectories for typical, strong-improvement, and difficult samples. Bicubic interpolation, outputs from $T=0$ to $T=4$, and the ground truth are shown using identical spatial crops.}
    \label{fig:progressive_trajectory}
\end{figure*}

\begin{figure*}[t]
    \centering
    \includegraphics[width=\textwidth]{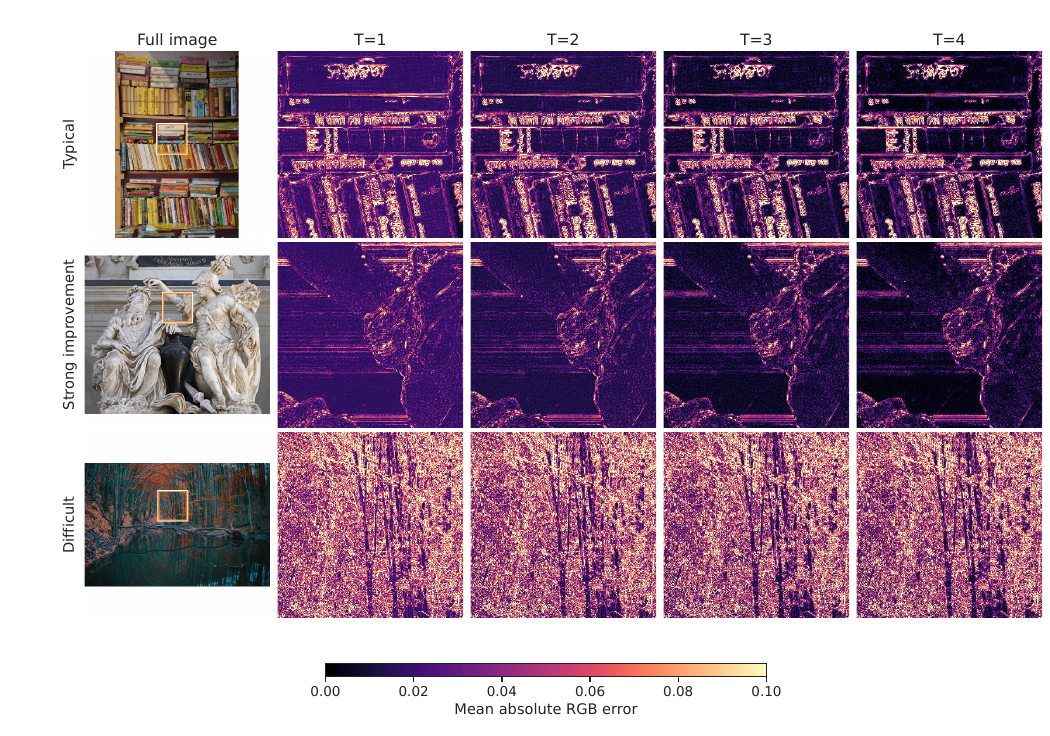}
    \caption{Progressive absolute-error maps corresponding exactly to the aligned reconstruction crops. All panels share the same error range and color mapping.}
    \label{fig:progressive_error_maps}
\end{figure*}

\subsection{Qualitative Thought Trajectory}

Fig.~\ref{fig:progressive_trajectory} shows the three representative reconstruction trajectories across the learned horizon. Fig.~\ref{fig:progressive_error_maps} shows the corresponding absolute-error maps under one shared display range and color mapping. These layouts preserve panel alignment from bicubic interpolation through $T=0$--$4$ and the ground truth.

\subsection{Current Boundary}

The current model uses a fixed four-tick training horizon, and extrapolation beyond this horizon is not guaranteed. The preliminary sweep is a feasibility measurement rather than a complete SOTA benchmark. Parameter-matched recurrent baselines, broader SR benchmarks, and verified evaluation details remain ongoing; no result from those studies is implied here.

\section{Discussion and Current Limitations}
\label{sec:discussion}

\subsection{Why Super-Resolution Is the First Instance}

SR makes the sparse-thought--dense-output mismatch directly observable. It requires a dense output, reacts visibly to lost spatial information, and permits the reconstruction at every thought tick to be measured with the same full-reference metrics. These properties make it a stringent diagnostic for whether compact thought can coordinate a persistent visual carrier.

\subsection{General Architectural Scope}

The DQ-CTM interface is not tied mathematically to RGB reconstruction. A persistent dense carrier could, in principle, be decoded into segmentation logits, a depth field, or optical flow. These are potential extensions only: none has been evaluated in the current V1, and their task-specific state, loss, topology, and stopping behavior remain open design questions.

\subsection{Current Limitations}

The evidence is limited to SR under a fixed four-tick training horizon. The present window flattening retains tokens but may not explicitly model the full two-dimensional topology during every thought operation. Adaptive stopping is not implemented, and no claim is made that quality improves indefinitely when additional ticks are applied. Evaluation details and recurrent comparisons must also be completed before strong empirical conclusions are appropriate. A concise future direction is to investigate whether Mamba-style spatial scanning can expose two-dimensional structure more explicitly without discarding the dense carrier.

\section{Conclusion}
\label{sec:conclusion}

This V1 report frames the sparse-thought--dense-output mismatch and describes a Dense-Query Continuous Thought Machine for addressing it. The central result is that a compact continuous-thought process can be coupled with a persistent dense visual carrier to produce a progressively refined high-resolution output. Its defining interface is a structured low-rank, parameter-efficient compact-to-dense transformation that converts each window-level CTM summary into distinct queries aligned with all positions in that window. In ThinkSR, local visual tokens remain available across shared thought ticks, synchronization-derived state governs dense updates, and reconstruction can be inspected at each point of the learned unroll. The preliminary four-tick sweep supports feasibility within that horizon, while leaving broader benchmarks, matched recurrent controls, implementation-level interface verification, and computation beyond the trained horizon unresolved. The formulation establishes a focused basis for studying continuous thought in dense vision without treating potential extensions as completed results.

{\small
\bibliographystyle{required_style_files/ieeenat_fullname}
\bibliography{references}
}

\clearpage
\appendix
\onecolumn
\setcounter{figure}{0}
\renewcommand{\thefigure}{S\arabic{figure}}

\section{Supplementary Attention Visualization}
\label{sec:supp_attention}

\begin{figure}[h]
    \centering
    \includegraphics[width=\textwidth]{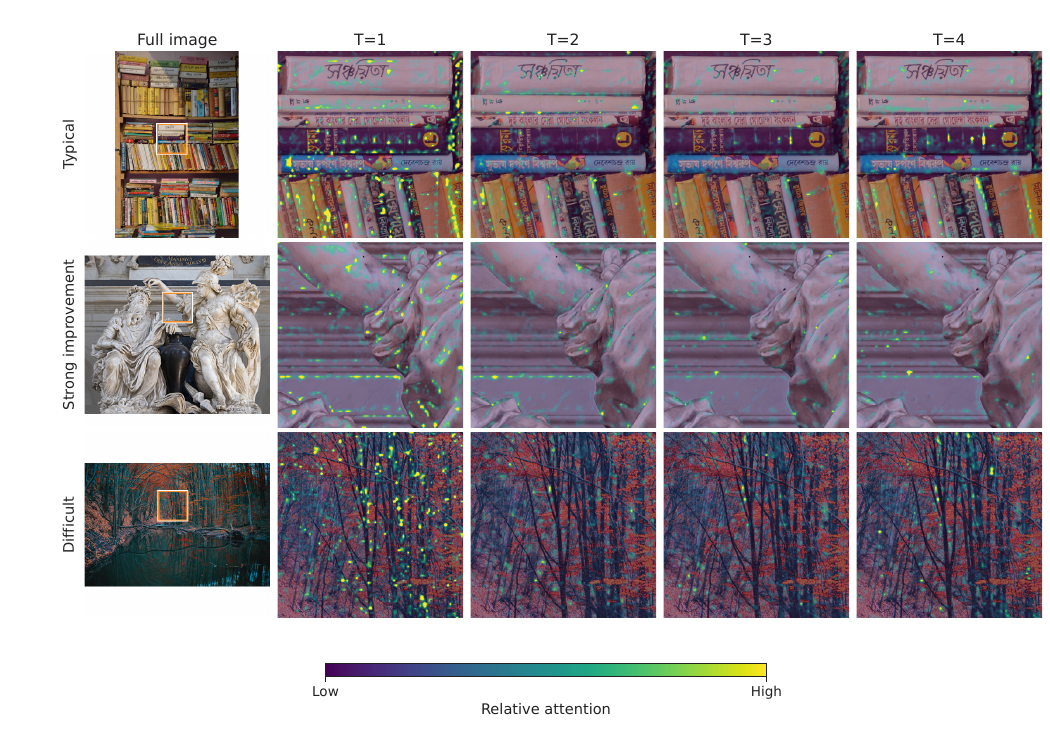}
    \caption{Relative attention overlays across thought ticks $T=1$--$4$ for the typical, strong-improvement, and difficult samples. Each row uses a fixed spatial crop, and all panels follow the same displayed relative-attention scale.}
    \label{fig:supp_attention_overlay}
\end{figure}

Fig.~\ref{fig:supp_attention_overlay} provides a qualitative view of how the supplied relative-attention visualization changes within the learned thought horizon. The crop for each sample is fixed across ticks. This diagnostic does not alter or recompute the reported reconstruction metrics, and the displayed intensity should not be interpreted as a causal attribution score.

\end{document}